\documentclass[runningheads]{llncs}
\usepackage[T1]{fontenc}
\usepackage{graphicx,verbatim}
\usepackage{booktabs}
\usepackage{amsmath}
\usepackage{amssymb}
\usepackage{comment}
\usepackage{float}

\begin{document}
\title{ProtoCLIP: Prototype-Aligned Latent Refinement for Robust Zero-Shot Chest X-Ray Classification}
\titlerunning{ProtoCLIP: Prototype-Aligned Zero-Shot Classification} 
%

\author{Florian Kittler, Sheethal Bhat, Andreas Maier}  
\authorrunning{Anonymized Author et al.}
\institute{Friedrich-Alexander University Erlangen-Nuremberg, Erlangen 91054, Germany \\
    \email{\{florian.flo.kittler, sheethal.bhat, andreas.maier\}@fau.de}}

\maketitle              
\begin{abstract}
Zero-shot vision–language models (VLMs) have shown prom\-ise for chest radiograph classification, but their performance is often limited by confounding label co-occurrence, long-tail class imbalance, and transfer instability under domain shift. We propose ProtoCLIP, a refinement strategy for CLIP-style VLMs that improves zero-shot discrimination through targeted data curation and distilled anchor alignment. Specifically, we construct pathology-focused training subsets with curated negative samples to reduce co-occurrence bias. We also introduce a representation-preserving distillation objective to stabilize adaptation while maintaining semantic structure and improving discrimination of clinically relevant co-occuring pathologies.
Evaluated on an unseen dataset VinDr-CXR, ProtoCLIP improves AUC by 2-10 percentage points over a strong CLIP-based baseline across multiple findings. For pneumothorax specifically, ProtoCLIP achieves a state-of-the-art AUC of 0.94. 
These results demonstrate that anchor-guided refinement, coupled with curated supervision and controlled adaptation, can mitigate common zero-shot transfer failures in medical VLMs without requiring large-scale retraining. 



\end{abstract}
\section{Introduction}

Chest radiography (CXR) is among the most frequently acquired medical imaging studies and remains a first-line modality for diagnosing and monitoring in a wide range of cardiopulmonary conditions \cite{ref_jones2021cxr_ml}. With rising imaging volumes and increasing clinical workload \cite{ref_alexander2022_workload}, scalable decision support tools are becoming essential. In time-critical scenarios such as suspected pneumothorax, where delayed recognition can lead to rapid deterioration, highly sensitive and reliable automated assistance is particularly important \cite{ref_roberts2014_tensionptx}.

Recent advances in vision–language models (VLMs), supported by large-scale CXR datasets such as MIMIC-CXR \cite{ref_johnson2019mimiccxr}, have accelerated data-driven radiograph interpretation \cite{ref_tiu2022CheXZero,ref_wu2023medklip}. These methods implement contrastive image–text learning that aligns radiographs with associated reports. This enables transferable representations that support zero-shot and prompt-based classification with minimal task-specific annotation \cite{ref_radford2021clip}. Radiology-specific adaptations, including CheXZero \cite{ref_tiu2022CheXZero}, GLORIA \cite{ref_huang2021gloria}, BioViL(T) \cite{ref_bannur2023biovilt}, CXR-CLIP \cite{ref_you2023cxrclip}, and MedKLIP \cite{ref_wu2023medklip}, leverage report-derived text embeddings to reduce reliance on curated labels and improve generalization. Beyond pre-training, refinement strategies such as prompt tuning, adapter-based updates, and concept-level alignment have been proposed to better match downstream clinical targets without full retraining \cite{ref_zhou2021coop,ref_zhang2022tipadapter}. However, these approaches primarily adjust decision boundaries rather than explicitly restructuring the embedding space to improve pathology separability.

Despite these advances, performance remains inconsistent for less prevalent and time-critical findings such as pneumothorax~\cite{ref_thian2021ptxexternal,ref_bhatbvm}. A key challenge is label co-occurrence and long-tailed prevalence in large-scale datasets, which can entangle pathology representations in the shared embedding space. When multiple findings frequently appear together, their visual–semantic features may cluster closely, reducing class separability and limiting zero-shot discrimination for underrepresented conditions. This lack of structured separation may hinder reliable deployment in high-sensitivity clinical scenarios.

To address this limitation, we propose ProtoCLIP, a pre-training refinement framework designed to improve pathology separability in the joint image–text embedding space. Our method performs a second-stage refinement of a CheXZero-initialized VL backbone, explicitly restructuring the joint embedding space through prototype anchoring and distillation. We optimize a dual-objective loss combining binary cross-entropy (BCE) alignment to these anchors with a feature distillation term that preserves the pretrained representation geometry. By anchoring image embeddings to stable semantic prototypes while constraining destructive drift, ProtoCLIP promotes more discriminative and transferable representations under distribution shift. We assess generalization on the external VinDr-CXR dataset using zero-shot AUC scores.

\subsubsection{Main Contributions:} 
(1) We introduce ProtoCLIP, a prototype-based refinement method that improves pathology separation in the joint embedding space of pre-trained CXR VL models.
(2) We curate a focused training subset to enhance label reliability and optimize a dual-objective learning scheme combining anchor-based alignment with teacher–student feature distillation.
(3) We perform out-of-distribution evaluations on VinDr-CXR, where ProtoCLIP achieves state-of-the-art (SOTA) zero-shot performance for pneumothorax and other selected pathologies. ProtoCLIP yields consistent gains across additional pathologies, including findings not explicitly prioritized during training. 

\section{Materials}

\subsubsection{Dataset description:}
MIMIC-CXR \cite{ref_johnson2019mimiccxr} is a large-scale chest radiograph dataset containing over 370,000 images from more than 65,000 patients, paired with free-text radiology reports and report-derived labels. We use frontal chest radiographs and curate a subset to improve label purity and control class imbalance for refinement training. MIMIC-CXR labels are derived from CheXpert-style report annotations generated by the CheXpert labeler, which assigns study-level binary findings from radiology reports \cite{ref_johnson2019mimiccxr}. 
VinDr-CXR \cite{ref_nguyen2022vindrcxr} is an external chest radiograph dataset comprising 15,000 training and 3,000 test studies collected from two hospitals. It provides 28 annotated findings, with test labels established by multi-radiologist consensus. We use the VinDr-CXR test set exclusively for out-of-distribution evaluation.

\subsubsection{Data preparation:}

To improve embedding separability for the target finding (e.g. pneumothorax), we curate a refined training subset from MIMIC-CXR \cite{ref_johnson2019mimiccxr} using the NLP generated labels \cite{miccai_ltcxr}. The dataset is divided into a (i) \emph{target} cohort $\mathcal{D}_t$ containing all pneumothorax-positive images (with multiple findings per image) and a (ii) \emph{background} cohort $\mathcal{D}_b$ constructed from single-pathology examples, where exactly one non-target finding is present. These are mathematically depicted below as,  
\begin{equation}
\mathcal{D}_t = \{x \in \mathcal{D} : y_t(x)=1\},
\label{eq:Dt}
\end{equation}
indicating all images labeled with the target finding $t$, irrespective of additional co-occuring pathologies. 
Let $\mathcal{B}$ be the set of all findings other than target, then for each $b\in\mathcal{B}$, we construct single-pathology background subset as,
\begin{equation}
 {\mathcal{D}}_b =
\{x \in \mathcal{D} : y_{b}(x)=1,\, y_{t}(x)=0, y_c(x)=0\  \forall c \in \mathcal{B}, c\neq b\}.
\label{eq:Db_pure}
\end{equation}
To mitigate class imbalance, each $\mathcal{D}_b$ is capped at randomly sampled $K=4000$ samples. The final curated dataset is, 
\begin{equation}
\mathcal{D}_{\mathrm{final}} = \mathcal{D}_t \cup \bigcup_{b\in\mathcal{B}} {\mathcal{D}}_b.
\label{eq:D_final}
\end{equation}

\section{Method}
\begin{figure}[ht]
\includegraphics[width=\textwidth]{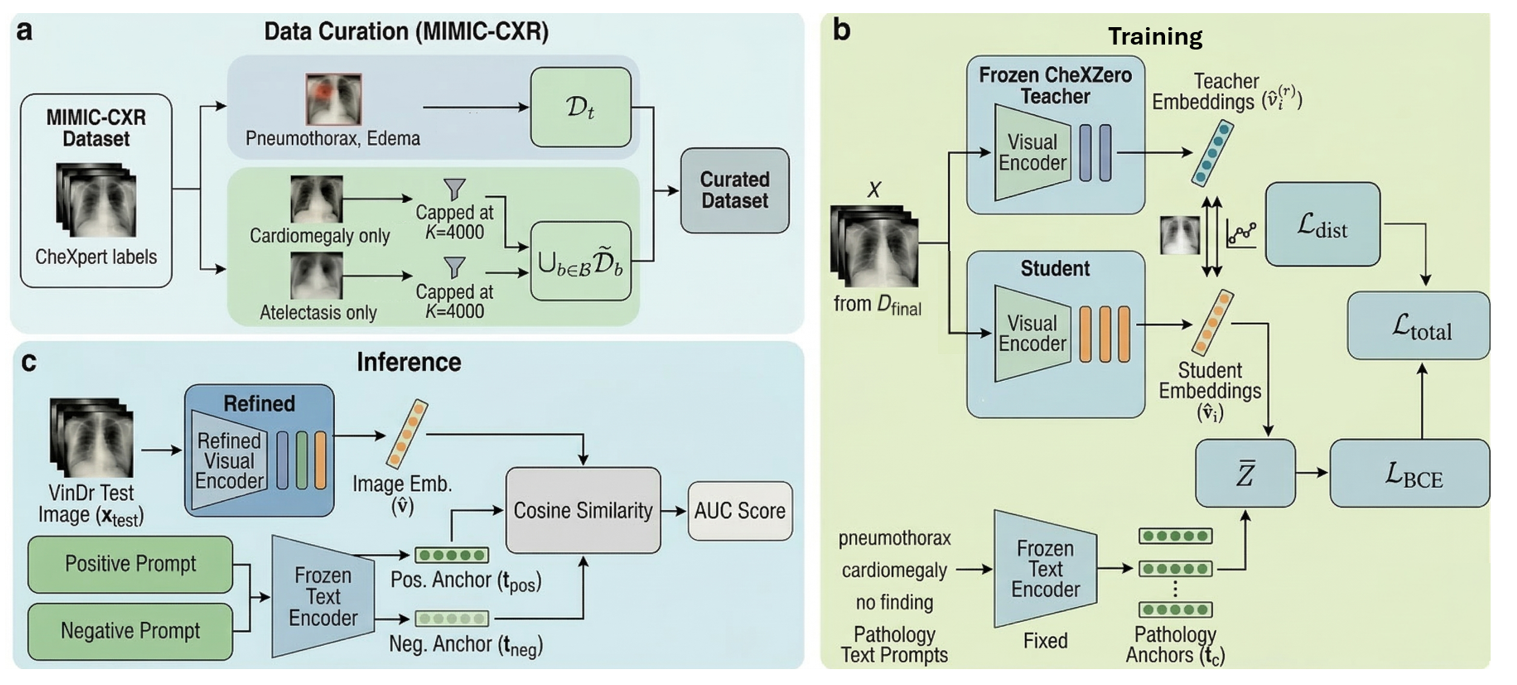}
\caption{\textbf{Overview of the ProtoCLIP method.} \textbf{a}, Data curation from MIMIC-CXR includes a set of pneumothorax-positive images $\mathcal{D}_t$, and single-pathology background examples ${\mathcal{D}}_b$. \textbf{b}, ProtoCLIP training schema depicts a trainable student visual encoder, refined using a dual-objective loss. $\mathcal{L}_{\mathrm{dist}}$ performs feature distillation from a frozen CheXZero teacher and $\mathcal{L}_{\mathrm{BCE}}$ aligns student image embeddings with frozen text-defined pathology anchors $t_c$, using one-hot targets  from the $\mathcal{D}_{\mathrm{final}}$.(\textbf{c}) At inference, the student visual encoder processes test image $\mathbf{x}_{\text{test}}$ to generate image embeddings $\hat{\mathbf{v}}$. The final classification score is determined by calculating the cosine similarity between $\hat{\mathbf{v}}$ and positive $\mathbf{t}_{\text{pos}}$ vs. negative $\mathbf{t}_{\text{neg}}$ text anchors.} \label{fig1}
\end{figure}

We adopt CheXZero as the underlying VL backbone and refine its joint embedding space using the curated training subset described above. ProtoCLIP introduces a dual-objective optimization scheme consisting of (i) a teacher–student feature distillation loss that preserves the pretrained representation structure and (ii) alignment to frozen text-defined pathology anchors via a BCE objective as depicted in Figure~\ref{fig1}. 



\subsubsection{Feature Distillation:}

To stabilize refinement, we incorporate feature-level knowledge distillation from a frozen CheXZero teacher \cite{ref_hinton2015distill}. This regularizes the student encoder, ensures the representation geometry is retained, and limits destructive drift in the embedding space. 
For each image $x$ in a batch $B$, let $\hat{\mathbf v}=f_\theta(x)$ and
$\hat{\mathbf v}^{(T)}=f_{\theta^{(T)}}(x)$ be $\ell_2$-normalized student and teacher embeddings.
We use the mean cosine distance to compute the distillation loss as,
\begin{equation}
\mathcal{L}_{\mathrm{dist}}
= \frac{1}{B}\sum_{i=1}^{B}\Bigl(1-\cos(\hat{\mathbf v}_i,\hat{\mathbf v_i}^{(T)})\Bigr).
\label{eq:L_dist}
\end{equation}


\begin{table}[ht]
\centering
\footnotesize
\caption{Zero-shot classification comparison of ProtoCLIP with current SOTA methods for pneumothorax (18/3000) on VinDR-CXR dataset \cite{ref_nguyen2022vindrcxr}. Sensitivity fixed at ~95\%  to minimize False Negatives in a clinical setting \cite{ref_bhat2026clinicalusability}. ProtoCLIP threshold $\gamma = 0.03$ \& CheXZero $\gamma = -0.002$.  ProtoCLIP AUC achieves $\pm 0.0028$ (SD), 95\% CI: $[0.9368,\,0.9438]$. Sensitivity, specificity, precision and F1 only compared to baseline.}
\label{tab:threshold_metrics_pneumo}
\begin{tabular*}{\textwidth}{@{\extracolsep{\fill}} l c c c c c}
\toprule
\textbf{Model} & \textbf{AUC} & \textbf{Sensitivity} & \textbf{Specificity} & \textbf{Precision} & \textbf{F1-score} \\
\midrule
Gloria \cite{ref_huang2021gloria}        & 0.68 & - & - & - & -\\
BioVil-T \cite{ref_bannur2023biovilt}   & 0.90 & - & - & - & -\\
CXR-CML \cite{ref_cxrcml}       & 0.51 & - & - & - & -\\
MedKLIP \cite{ref_wu2023medklip}        & 0.93 & - & - & - & -\\ 
CheXZero \cite{ref_tiu2022CheXZero} & 0.85 & 0.94 & 0.71 & 0.02 & 0.04 \\
ProtoCLIP (Ours) & \textbf{0.94} & 0.94 & \textbf{0.84} & \textbf{0.03} & \textbf{0.07} \\
\bottomrule
\end{tabular*}
\end{table}

\subsubsection{Anchor-Based Prototype Alignment:}

We align image embeddings to fixed pathology prototypes using a standard BCE objective. During refinement, only the final two layers of the student visual encoder are updated, while earlier visual layers and the text encoder remain frozen. The frozen text encoder produces a fixed set of $C$ anchor embeddings $\{\mathbf{t}_c \mid c \in \{1, \dots, C\}\}$ in the shared embedding space. For a batch of $B$ images, the student generates image embeddings whose similarity to each anchor forms a logit matrix $\mathbf{Z}\in\mathbb{R}^{B\times C}$ where $\mathbf{Z}_{c} = \tau \cdot \cos(\hat{\mathbf{v}}_i, \mathbf{t}_c)$ and $\tau$ is a temperature scaling parameter. Since each curated image belongs to exactly one bucket, supervision is represented by one-hot targets $\mathbf{Y}\in\{{0,1}\}^{B\times C}$. The standard BCE loss is computed using $\mathbf{Z}$ and $\mathbf{Y}$. We use BCE (rather than softmax CE) to keep compatibility with multi-label settings and to avoid forcing relative competition across anchors.
The final loss is a weighted sum of both denoted as,
\begin{equation}
\mathcal{L} \;=\; \mathcal{L}_{\mathrm{BCE}} \;+\; \lambda\, \mathcal{L}_{\mathrm{dist}},
\label{eq:loss_total}
\end{equation}
where $\lambda$ controls the trade-off between new signals for improved clustering and old representations to prevent catastrophic forgetting.

\section{Results}

\begin{table}[ht]
\caption{AUC comparison for the two most (Pleural Effusion \& Atelectasis) and the two least co-occuring (Pneumonia \& Consolidation) background classes when pneumothorax is the target class.}
\label{tab:cooccurrence_auc_simple}
\centering
\footnotesize
\begin{tabular*}{\textwidth}{@{\extracolsep{\fill}} l c c}
\toprule
\textbf{Pathology} & \textbf{CheXZero} \cite{ref_tiu2022CheXZero} & \textbf{ProtoCLIP (Ours)} \\
\midrule
Pleural Effusion & 0.9550 & \textbf{0.9601} \\
Atelectasis      & 0.6884 & \textbf{0.7658} \\
\midrule
Pneumonia        & \textbf{0.8922} & 0.8906 \\
Consolidation    & \textbf{0.9235} & 0.9214 \\
\bottomrule
\end{tabular*}
\end{table}

\subsection{Experimental Setup:}

We use the ViT-B/32 backbone from CheXZero \cite{ref_tiu2022CheXZero} and train on a single NVIDIA Tesla V100 (32GB). Images are resized to $224\times224$. The model is implemented in PyTorch 2.4.1 with a learning rate of $1\times10^{-4}$ and batch size 128. We adopt the standard CLIP temperature $\tau=0.07$ \cite{ref_radford2021clip} and set the distillation weight $\lambda=1.0$. Main results for pneumothorax are reported as the mean over five independent runs using random seeds and other pathologies are evaluated using a single checkpoint. Fixed text anchors use the templates "\{{pathology}\}" and “indicating \{pathology\}”.


\subsection{Comparison to SOTA:}

\begin{table}[ht]

\caption{Zero-shot AUC comparison of different target pathologies between current SOTA and ProtoCLIP. }
\label{tab:per_pathology_auc}
\centering
\footnotesize
\begin{tabular*}{\textwidth}{@{\extracolsep{\fill}} l c c c c c}
\toprule
\textbf{Pathology} & \textbf{Gloria} \cite{ref_huang2021gloria} & \textbf{BioVil-T} \cite{ref_bannur2023biovilt} & \textbf{CheXZero} \cite{ref_tiu2022CheXZero} & \textbf{MedKLIP} \cite{ref_wu2023medklip} & \textbf{ProtoCLIP} \\
\midrule
Lung opacity      & 0.63 & 0.79 & 0.79 & 0.83 & \textbf{0.84} \\
Atelectasis       & 0.61 & 0.67 & 0.72 & \textbf{0.87} & 0.82 \\
Pleural effusion  & 0.67 & 0.95 & 0.95 & \textbf{0.96} & \textbf{0.96} \\
\bottomrule
\end{tabular*}
\end{table}

Table~\ref{tab:threshold_metrics_pneumo} compares zero-shot pneumothorax classification performance of ProtoCLIP vs. various VL methods.
ProtoCLIP achieves the highest AUC ($0.94$), outperforming CheXZero (0.85) and slightly surpassing MedKLIP (0.93). Performance is stable across runs (±0.0028 SD; 95\% CI [0.9368, 0.9438]) when compared to CheXZero with SD $\pm 0.09$ \cite{ref_bhatbvm}. 
At a clinically relevant operating point \cite{ref_bhat2026clinicalusability}, ProtoCLIP achieves substantially higher specificity than CheXZero ($0.8390$ vs.\ $0.7123$), corresponding to fewer false positives at the same sensitivity. This also leads to improved precision and F1-score (0.0342/0.0660 vs.\ 0.0194/0.0381), indicating more reliable positive predictions in a clinically relevant high-sensitivity setting.

Table~\ref{tab:cooccurrence_auc_simple} reports AUC scores for selected background findings under pneumothorax targeted training. Compared to CheXZero, ProtoCLIP improves performance on pleural effusion and atelectasis (common co-occurring confounders) suggesting better disentanglement of overlapping visual patterns. The gain for atelectasis is particularly notable (+0.08 AUC). In contrast, AUC for pneumonia and consolidation changes only marginally, indicating that the refinement mainly benefits frequent co-occurring findings while preserving performance on less related classes.

Table 3 depicts the AUC comparison across different pathologies as target class, where ProtoCLIP attains the best performance for Lung Opacity (0.84) and matches the top result for Pleural Effusion (0.96), demonstrating that the proposed refinement generalizes beyond pneumothorax. For Atelectasis, ProtoCLIP substantially improves over CheXZero (0.82 vs.\ 0.72) but remains below MedKLIP (0.87), indicating pathology-dependent gains potentially influenced by label prevalence and co-occurrence structure. Overall, anchor-guided refinement consistently enhances zero-shot discrimination across several findings, though it does not uniformly surpass specialized baselines.

Although ProtoCLIP performs well, its main limitation is that it is currently optimized for a single target pathology rather than serving as a general-purpose refinement method. Nevertheless, we find that improvements extend to several background classes for a given target class suggesting potential broader applicability. Additionally, further validation on independent and out-of-distribution cohorts is necessary to establish generalizability. CheXpert~\cite{ref_irvin2019chexpert} was excluded because pneumothorax is not in the official Competition 5 benchmark and its very low prevalence precludes statistically reliable evaluation.


\subsubsection{Visualization:}

\begin{figure}[ht]
\centering
\includegraphics[width=0.8\textwidth]{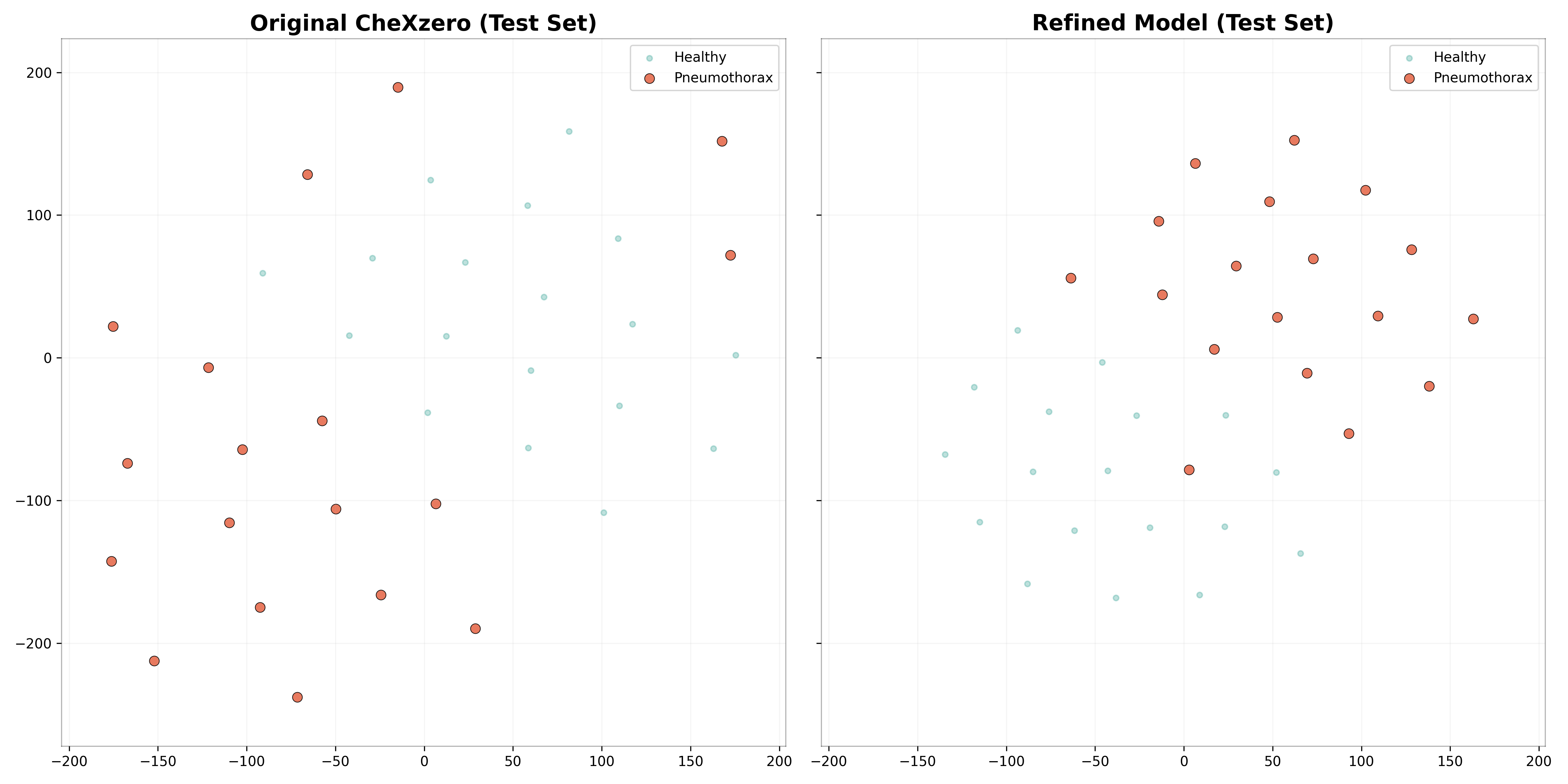}
\caption{Comparison of TSNE clustering of pneumothorax (n=18) vs. healthy (n=18) samples between CheXZero \cite{ref_tiu2022CheXZero} and ProtoCLIP.} \label{fig:tsne}
\end{figure}

To qualitatively assess the effect of the proposed refinement, we analyze embedding distributions and attention maps for the target pneumothorax.
\paragraph{TSNE:}
Due to the large class imbalance for pneumothorax, the tsne visualization for the entire dataset was inconclusive. However, Fig.~\ref{fig:tsne} depicts  a qualitative analysis over equal number of True Positives (TP) for pneumothorax cases and healthy findings. This shows a modest improvement in inter and intra-class clustering.

\paragraph{Attention maps:}

\begin{figure}[ht]
\centering
\includegraphics[width=1\textwidth]{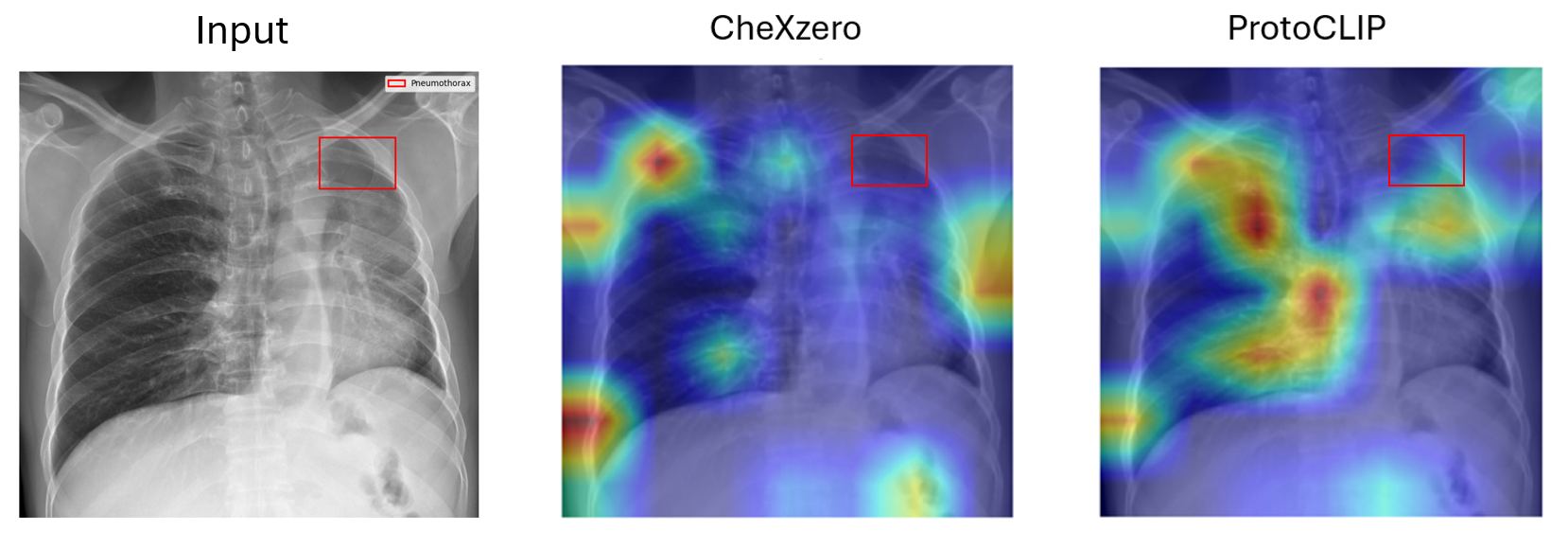}
\caption{Attention map comparison of TP pneumothorax from VinDR with bounding box annotation.} \label{fig:attn_maps}
\end{figure}

Zero-shot pneumothorax classification with VLMs does not inherently guarantee anatomically meaningful attention maps. As illustrated in Fig.~\ref{fig:attn_maps}, ProtoCLIP yields attention that is more consistently concentrated within the lung fields compared to the CheXZero baseline, with partial alignment to the ground-truth bounding box. This qualitative improvement is consistent with the observed gains in classification performance over the baseline.

\subsubsection{Ablation studies:}

\begin{table}[ht]
\caption{AUC comparison of different hyperparameter settings, when pneumothorax is the target class, across different batch size $bs$ and distillation weights $\lambda$, as well as increased number and variety of text anchors.   }
\label{tab:ablation_auc}
\centering
\footnotesize
\begin{tabular*}{\textwidth}{@{\extracolsep{\fill}} l c}
\toprule
\textbf{Hyperparameter setting} & \textbf{AUC Score} \\
\midrule
Distill $\lambda$ = 0, $bs$ = 128 ($\mathcal{L}_{dist}=0$)         & 0.9237 \\
Distill $\lambda$ = 1, bs = 128        & \textbf{0.9426} \\
Distill $\lambda$ = 10, bs = 128         & 0.9378 \\
Distill $\lambda$ = 1, bs = 64 & 0.9331 \\
Distill $\lambda$ = 1, bs = 128 & 0.9426 \\
Distill $\lambda$ = 1, bs = 256             & 0.9352 \\
Distill $\lambda$ = 1, bs = 128, multiple complex text anchors & 0.9397 \\
\midrule
Best performing model, bs = 128, $\lambda$ = 1,  single text anchors & \textbf{0.9426} \\
\bottomrule
\end{tabular*}
\end{table}

We perform ablation studies over multiple hyper-parameters and settings. As shown in Table~\ref{tab:ablation_auc}, a batch size of 128 achieves the best performance, while smaller (64) and larger (256) batches slightly degrade results. Increasing the distillation weight to $\lambda=10$ provides no benefit, indicating over-regularization, whereas removing distillation ($\lambda=0$) yields the worst performance, particularly for low co-occurrence classes, highlighting its importance for representation stability. Increasing the number and diversity of text anchor templates does not lead to consistent gains. The optimal configuration therefore uses batch size 128, $\lambda=1$, and simple text templates.

\section{Conclusion}

We introduced ProtoCLIP, a refinement strategy for CLIP-style VLMs that improves zero-shot performance in chest radiography through targeted data curation and distilled anchor alignment. Evaluation on VinDr-CXR shows that anchor-guided refinement combined with representation-preserving regularization mitigates common transfer limitations and enhances robustness under domain shift. Future work will focus on automating data curation, validating the framework on additional datasets, and extending the approach toward a unified model covering multiple pathologies.

\newpage
%
%
%

\begin{thebibliography}{15}
\bibitem{ref_jones2021cxr_ml}
Jones, C.M., Buchlak, Q.D., Oakden-Rayner, L., Milne, M., Seah, J., Esmaili, N., Hachey, B.:
Chest radiographs and machine learning -- Past, present and future.
Journal of Medical Imaging and Radiation Oncology \textbf{65}(5), 538--544 (2021). \doi{10.1111/1754-9485.13274}

\bibitem{ref_alexander2022_workload}
Alexander, R., Waite, S., Bruno, M.A., Krupinski, E.A., Berlin, L., Macknik, S., Martinez-Conde, S.:
Mandating limits on workload, duty, and speed in radiology.
Radiology \textbf{304}(2), 274--282 (2022). \doi{10.1148/radiol.212631}

\bibitem{ref_roberts2014_tensionptx}
Roberts, D.J., Leigh-Smith, S., Faris, P.D., et al.:
Clinical manifestations of tension pneumothorax: protocol for a systematic review and meta-analysis.
Systematic Reviews \textbf{3}, 3 (2014). \doi{10.1186/2046-4053-3-3}

\bibitem{ref_johnson2019mimiccxr} 
Johnson, A.E.W., Pollard, T.J., Berkowitz, S.J., Greenbaum, N.R., Lungren, M.P., Deng, C.Y., Mark, R.G., Horng, S.:
MIMIC-CXR, a de-identified publicly available database of chest radiographs with free-text reports.
Scientific Data \textbf{6}(1), 317 (2019). \doi{10.1038/s41597-019-0322-0}

\bibitem{ref_tiu2022CheXZero} 
Tiu, E., Talius, E., Patel, P., Langlotz, C.P., Ng, A.Y., Rajpurkar, P.:
Expert-level detection of pathologies from unannotated chest X-ray images via self-supervised learning.
Nature Biomedical Engineering \textbf{6}(12), 1399--1406 (2022). \doi{10.1038/s41551-022-00936-9}

\bibitem{ref_cxrcml}
Madhipati, R., Maier, A.:
CXR-CML: Improved zero-shot classification of long-tailed multi-label diseases in Chest X-Rays. 
In: International Conference on Medical Image Computing and Computer-Assisted Intervention (pp. 119-129). Cham: Springer Nature Switzerland 2025.

\bibitem{ref_wu2023medklip} 
Wu, C., Zhang, X., Zhang, Y., Wang, Y., et al.:
MedKLIP: Medical Knowledge Enhanced Language-Image Pre-Training for X-ray Diagnosis.
In: Proceedings of the IEEE/CVF International Conference on Computer Vision (ICCV), pp. 1--12 (2023). \doi{10.1109/ICCV51070.2023.01954}

\bibitem{ref_du2022vlpsurvey}
Du, Y., Liu, Z., Li, J., Zhao, W.X.:
A Survey of Vision-Language Pre-Trained Models.
In: Proceedings of the Thirty-First International Joint Conference on Artificial Intelligence (IJCAI),
pp.~5436--5443 (2022). \doi{10.24963/ijcai.2022/762}

\bibitem{ref_radford2021clip} 
Radford, A., Kim, J.W., Hallacy, C., Ramesh, A., Goh, G., Agarwal, S., Sastry, G., Askell, A., Mishkin, P., Clark, J., Krueger, G., Sutskever, I.:
Learning Transferable Visual Models From Natural Language Supervision.
In: Proceedings of the 38th International Conference on Machine Learning (ICML),
PMLR, vol.~139, pp.~8748--8763 (2021). \doi{10.48550/arXiv.2103.00020}

\bibitem{ref_huang2021gloria} 
Huang, S.-C., Shen, L., Lungren, M.P., Yeung, S.:
GLoRIA: A Multimodal Global-Local Representation Learning Framework for Label-Efficient Medical Image Recognition.
In: Proceedings of the IEEE/CVF International Conference on Computer Vision (ICCV), pp. 3942--3951 (2021)

\bibitem{ref_bannur2023biovilt} 
Bannur, S., Hyland, S., Liu, Q., P\'erez-Garc{\'\i}a, F., Ilse, M., Castro, D.C., Boecking, B., Sharma, H., Bouzid, K., Thieme, A., Schwaighofer, A., Wetscherek, M., Lungren, M.P., Nori, A., Alvarez-Valle, J., Oktay, O.:
Learning To Exploit Temporal Structure for Biomedical Vision-Language Processing.
In: Proceedings of the IEEE/CVF Conference on Computer Vision and Pattern Recognition (CVPR), pp. 15016--15027 (2023)

\bibitem{ref_zhou2021coop}
Zhou, K., Yang, J., Loy, C.C., Liu, Z.:
Learning to Prompt for Vision-Language Models.
arXiv preprint arXiv:2109.01134 (2021). \doi{10.48550/arXiv.2109.01134}

\bibitem{ref_zhang2022tipadapter}
Zhang, R., Wei, Z., Fang, R., Gao, P., Li, K., Dai, J., Qiao, Y., Li, H.:
Tip-Adapter: Training-Free Adaption of CLIP for Few-shot Classification.
In: European Conference on Computer Vision (ECCV), pp. 1--17 (2022).

\bibitem{ref_thian2021ptxexternal}
Thian, Y.L., et al.:
Deep Learning Systems for Pneumothorax Detection on Chest Radiographs: A Multicenter External Validation Study.
Radiology: Artificial Intelligence \textbf{3}(4), e200190 (2021). \doi{10.1148/ryai.2021200190}

\bibitem{ref_you2023cxrclip}
You, K., Gu, J., Ham, J., Park, B., Kim, J., Hong, E.K., Baek, W., Roh, B.:
CXR-CLIP: Toward Large Scale Chest X-ray Language-Image Pre-training.
In: Greenspan, H. et al. (eds.) MICCAI 2023, LNCS, vol.~14221, pp. 101--111.
Springer, Cham (2023). \doi{10.1007/978-3-031-43895-0_10}

\bibitem{ref_bishop2006prml}
Bishop, C.M.:
Pattern Recognition and Machine Learning.
Springer, New York (2006)

\bibitem{ref_hinton2015distill}
Hinton, G., Vinyals, O., Dean, J.:
Distilling the Knowledge in a Neural Network.
arXiv preprint arXiv:1503.02531 (2015). \doi{10.48550/arXiv.1503.02531}

\bibitem{ref_nguyen2022vindrcxr}
Nguyen, H.Q., Lam, K., Le, L.T., et al.:
VinDr-CXR: An open dataset of chest X-rays with radiologist's annotations.
Scientific Data \textbf{9}, 429 (2022). \doi{10.1038/s41597-022-01498-w}

\bibitem{ref_bhat2026clinicalusability}
Bhat, S., Maier, A.:
AUCReshaping: improved sensitivity at high-specificity.
In: Scientific Report (2023). Springer, \doi{https://doi.org/10.1038/s41598-023-48482-x}

\bibitem{ref_irvin2019chexpert}
Irvin, J., Rajpurkar, P., Ko, M., Yu, Y., Ciurea-Ilcus, S., Chute, C., Marklund, H., Haghgoo, B., Ball, R., Shpanskaya, K., Seekins, J., Mong, D.A., Halabi, S.S., Sandberg, J.K., Jones, R., Larson, D.B., Langlotz, C.P., Patel, B.N., Lungren, M.P., Ng, A.Y.:
CheXpert: A Large Chest Radiograph Dataset with Uncertainty Labels and Expert Comparison.
In: Proceedings of the AAAI Conference on Artificial Intelligence, \textbf{33}(01), 590--597 (2019). \doi{10.1609/aaai.v33i01.3301590}

\bibitem{ref_bhatbvm}
Bhat, S., Maier, A.:
Towards robust zero-shot chest x-ray classification: exploring data distribution bias in chest x-ray datasets. 
In: BVM Workshop (2025) (pp. 191-196). Wiesbaden: Springer Fachmedien Wiesbaden.

\bibitem{miccai_ltcxr}
Lin, M.,  Peng, Y.:
CXR-LT 2024: A MICCAI challenge on long-tailed, multi-label, and zero-shot disease classification from chest X-ray. Medical Image Analysis.
In: Medical Image Analysis, 2025, 103739.
\end{thebibliography}
%

\end{document}